\documentclass[a4paper,conference]{IEEEtran}
\pdfoutput=1

\usepackage{etoolbox}
\makeatletter
\patchcmd{\@makecaption}
  {\scshape}
  {}
  {}
  {}
\makeatother

\usepackage{tikz}
\usepackage{pgfplots}
\pgfplotsset{width=7.5cm,compat=1.12}
\usepgfplotslibrary{fillbetween}

\usepackage{hyperref}

\ifCLASSINFOpdf
\else
\fi
\begin{document}
%
\title{Traditional Chinese Synthetic Datasets Verified with Labeled Data for Scene Text Recognition}




%
\author{\IEEEauthorblockN{Yi-Chang Chen${^1}$,
Yu-Chuan Chang${^1}$,
Yen-Cheng Chang${^1}$ and
Yi-Ren Yeh${^2}$}
\\
\IEEEauthorblockA{$^1$E.SUN Financial Holding CO., LTD., Taiwan}
\IEEEauthorblockA{$^2$Department of Mathematics, National Kaohsiung Normal University, Taiwan}
\IEEEauthorblockA{E-MAIL: ycc.tw.email@gmail.com, chuanworks@gmail.com, timtimchang9666@gmail.com, yryeh@nknu.edu.tw}
}


\maketitle

\begin{abstract}
Scene text recognition (STR) has been widely studied in academia and industry. Training a text recognition model often requires a large amount of labeled data, but data labeling can be difficult, expensive, or time-consuming, especially for Traditional Chinese text recognition. To the best of our knowledge, public datasets for Traditional Chinese text recognition are lacking. This paper presents a framework for a Traditional Chinese synthetic data engine which aims to improve text recognition model performance. We generated over 20 million synthetic data and collected over 7,000 manually labeled data {\tt TC-STR 7k-word} as the benchmark\footnote{These two datasets are available at \url{https://github.com/GitYCC/traditional-chinese-text-recogn-dataset}}. Experimental results show that a text recognition model can achieve much better accuracy either by training from scratch with our generated synthetic data or by further fine-tuning with {\tt TC-STR 7k-word}.
\end{abstract}


%
\IEEEpeerreviewmaketitle

\section{Introduction}
\label{intro}

Scene text recognition is a challenging task due to the variety of text styles and backgrounds. Typically, a large amount of data is required to train a competitive recognition model. To the best of our knowledge, public datasets for Traditional Chinese text recognition are lacking, and the wide range of fonts and styles for Chinese characters causes additional difficulty for data collection, which is already difficult, expensive, and time-consuming.

Synthetic data is increasingly used to reduce the effort required for data collection and labeling for training deep learning models, especially in computer vision \cite{mjsynth,Ankush_2016_CVPR}. However, collecting a sufficient volume and variety of characters for Traditional Chinese text recognition is difficult. To overcome this problem, we generate synthetic scene text images with greater diversity.

The proposed synthetic data engine for Traditional Chinese scene text images begins by designing different attributes for generating text images, such as font rendering, background rendering, and text properties. Based on these different attributes, the synthetic word data generator imitates scene text from the real world to generate text images without human labeling. In addition to the synthetic data engine, we also create a real-world Traditional Chinese scene text dataset for evaluation. All text boxes and labels are cropped and annotated manually from real-world images.

In our experiments, we investigate the influence of different dataset sizes and the impacts of different attributes. We also show that the model pre-trained with our synthetic dataset outperforms one trained with the target dataset (if the target data are available), and that our synthetic data engine could significantly reduce the cost of human labeling and enhance text recognition performance.

\begin{figure*}[t]
\centerline{\includegraphics[width=18cm]{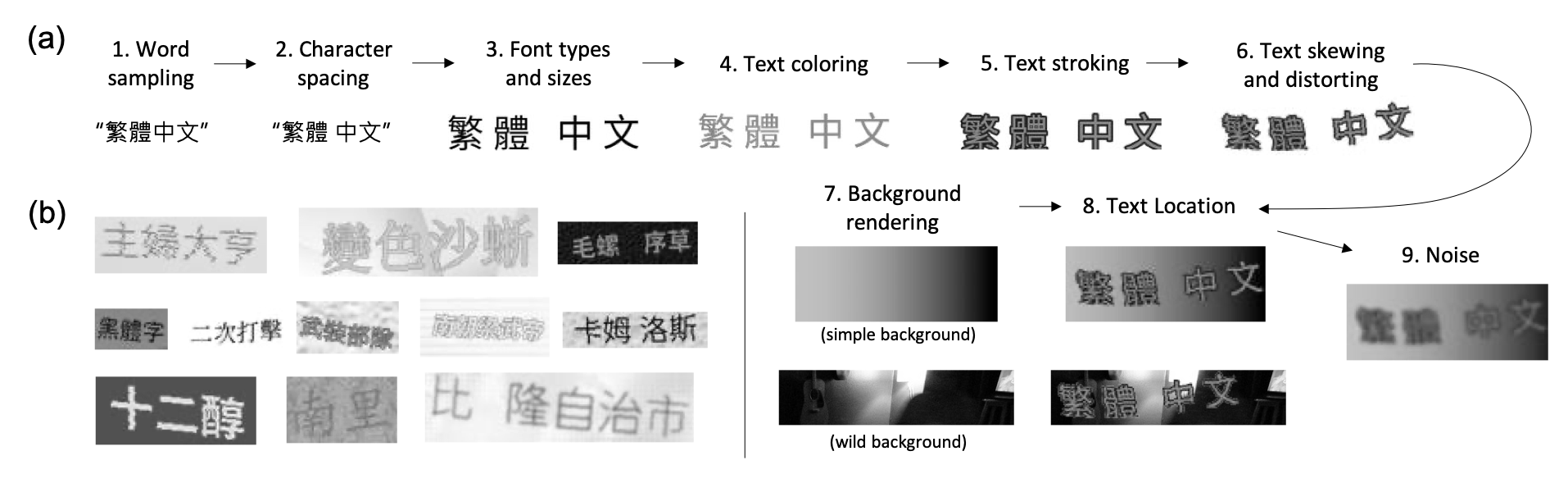}}
\caption{(a) The nine steps of our Traditional Chinese synthetic text engine. (b) Some randomly sampled examples generated from our synthetic text engine.}
\label{fig1}
\end{figure*}

\begin{figure}[t]
\centerline{\includegraphics[width=9cm]{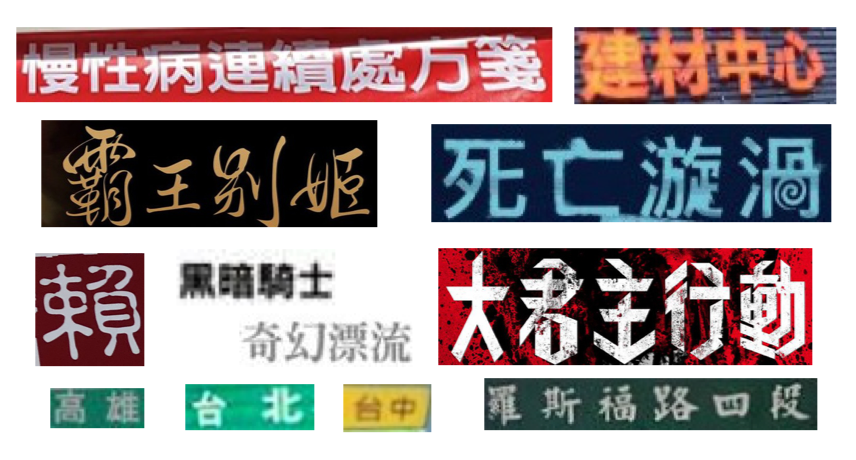}}
\caption{Some randomly sampled examples from the {\tt TC-STR 7k-word} dataset.}
\label{fig2}
\end{figure}

\section{Related Work} 
\label{related}

As mentioned in Section \ref{intro}, synthetic data have been widely used in training deep neural networks. \cite{mjsynth} proposed a popular synthetic dataset for real-world text recognition, the {\tt MJSynth} dataset, containing 8.9 million text images and 1,400 different fonts. The {\tt MJSynth} dataset is composed of three separate image layers: background, foreground, and optional shadow/border. Each text is synthesized with different font properties, such as kerning, weight, and underline. The {\tt SynthText} dataset \cite{synthtext} is another popular synthetic dataset used in text recognition, consisting of 5.5 million word images for which the synthetic text is blended with existing background images according to the local 3D scene geometry.

The most popular methods for recognition models use multi-stage pipelines \cite{b11}, including transformation, feature extraction, sequence modeling, and prediction stages. We use the thin-plate spline (TPS) transformation in the transformation stage. TPS is a variant of the spatial transformation network (STN) and has been shown to treat diverse aspect ratios of text lines \cite{b25, b06}. For the feature extraction stage, many CNN architectures have been proposed for the feature extraction. Based on the results in \cite{b11}, we choose ResNet as our feature extractor due to its superior performance. According to suggestions from \cite{b11}, we apply Bidirectional LSTM (BiLSTM) to catch contextualized information from the scene text images. In the last stage, Connectionist Temporal Classification (CTC) \cite{Graves_ICML_2006} brings more benefits for languages with large character sets, such as Traditional Chinese \cite{b20}. Departing from \cite{b11}, we used CTC instead of the encoder-decoder framework at the prediction stage. In a nutshell, we choose the combination TPS-ResNet-BiLSTM-CTC for Chinese scene text recognition in our experiments.

\section{Proposed Synthetic and Real-world Traditional Chinese Scene Text Datasets}
\label{synthetic}

\subsection{Traditional Chinese Synthetic Scene Text Engine} \label{text-generator}
Inspired by {\tt MJSynth} \cite{mjsynth}, {\tt SynthText} \cite{synthtext} and Belval/TextRecognitionDataGenerator\footnote{\url{https://github.com/Belval/TextRecognitionDataGenerator}}, we propose a framework for generating scene text images for Traditional Chinese. To produce synthetic text images similar to real-world ones, we use different kinds of mechanisms for rendering, as shown in Fig.~\ref{fig1}(a). The details of our data generating pipeline are as follows:

\begin{enumerate}
\item \textbf{Word sampling} -- In our synthetic scene text dataset, each synthesized text image is associated with a word that contains several characters as shown in Fig.~\ref{fig1}(b). To obtain a more diverse word set for word sampling, we extract words from two sources: Taiwan Ministry of Education dictionary\footnote{\url{https://language.moe.gov.tw/001/Upload/Files/site_content/M0001/respub/index.html}} and Wikipedia page titles. Our word set contains 1,076,764 words and 12,108 characters. 

On the other hand, to produce different appearances and backgrounds of the same word, we also repeatedly sample each word multiple times, each time applying different rendering tricks, such as font types, font sizes, font colors, stroking, skewing, distorting, simple and wild backgrounds, word location, and noise.

\item \textbf{Character spacing} -- It is common to have spaces between Chinese characters, especially in scene texts. To produce near-authentic Chinese scene texts, we randomly insert multiple spaces between characters within a word. As shown in the second step of Fig.~\ref{fig1}(a), we insert a space between the third and fourth characters where the location and the number of spaces are randomly determined.

\item \textbf{Font types and sizes} -- In our proposed synthetic dataset, we gathered 175 fonts with commercial-free authorization. The typefaces of those fonts include Gothic, Ming, Kai, Yuan, etc. For each synthetic image, one of the collected fonts is randomly selected for the rendering of font types. All fonts are sized from 20 to 50 points.

\item \textbf{Text coloring} -- In our image preprocessing, all the scene text images will be converted to grayscale before training the text recognition model. Thus, we simply fill the text with 14 different grayscale intensities. The hex color codes of our text coloring are \#000000, \#141414, \#282828, \#3C3C3C, \#505050, \#646464, \#787878, \#8C8C8C, \#A0A0A0, \#B4B4B4, \#C8C8C8, \#DCDCDC, \#F0F0F0, and  \#FFFFFF.

\item \textbf{Text stroking} -- To produce different text outline styles, we also assign different widths of text stroking. The widths are randomly selected from 0 to 3 points as shown in the fourth step of Fig.~\ref{fig1}(a).

\begin{table*}[t]
\caption{Results tested on {\tt TC-STR-test}. The first and second rows present recognition results of recognition models trained without our synthetic data. The third, fourth and fifth rows present recognition results of recognition models trained and validated with our synthetic data and {\tt TC-STR-Train} respectively. The last row presents the result of the model both trained and validated with our synthetic data.}
\begin{center}
\begin{tabular}{lr|lr|r}
\hline
\multicolumn{2}{c|}{\textbf{Training Data}} & \multicolumn{2}{c|}{\textbf{Validation Data}} & \multicolumn{1}{c}{\textbf{Test}} \\
\multicolumn{1}{c}{\textbf{name}} & \multicolumn{1}{c|}{\textbf{\# of data}} & \multicolumn{1}{c}{\textbf{name}} & \multicolumn{1}{c|}{\textbf{\# of data}} &  \multicolumn{1}{c}{\textbf{Accuracy}} \\
\hline
Subset of {\tt TC-STR-train} & 3,251 & Subset of {\tt TC-STR-train} & 586 & 4.10\% \\
Subset of {\tt TC-STR-train} + Augmentation \cite{data-aug} & 1,232,129 & Subset of {\tt TC-STR-train} & 586 & 8.82\% \\
\hline
Our Synthetic Data w/ Simple BGs & 1,076,764 & {\tt TC-STR-train} & 3,837 & 78.44\% \\
Our Synthetic Data w/ Simple BGs & 16,151,460 & {\tt TC-STR-train} & 3,837 & 84.11\% \\
Our Synthetic Data w/ Mixed BGs & 21,535,280 & {\tt TC-STR-train} & 3,837 & \textbf{84.75\%} \\
\hline
Our Synthetic Data w/ Mixed BGs & 21,535,280 & Our Synthetic Data w/ Mixed BGs & 6,000 & 83.51\% \\
\hline
\end{tabular}
\end{center}
\label{table-all-results}
\end{table*}

\item \textbf{Text skewing and distorting} -- 
Real-world scene texts are not often well-aligned horizontally due to appearance preferences or environmental constraints, such as the real-world examples in Fig.~\ref{fig2}. To simulate these properties, we skew the text after coloring/stroking and distort it vertically and horizontally. The sixth step of Fig.~\ref{fig1}(a) shows an example of our text skewing and distorting.

\item \textbf{Background rendering} -- There are two kinds of raw background images in background rendering. The first is simple background images retrieved from Google image search with specific queried keywords, such as slide background and texture. The other kind is wild background images, which are extracted from the {\tt COCO} \cite{coco_dataset} dataset. However, some images from the {\tt COCO} dataset contain texts. For example, the {\tt COCO-Text} \cite{coco_text} dataset selects images that contain texts from the {\tt COCO} dataset, and labels those images with locations and texts. Thus, to avoid extracting wild background images with texts, we excluded those images in the {\tt COCO} dataset by referencing the labels of the {\tt COCO-Text} \cite{coco_text} dataset. The examples of simple and wild backgrounds are respectively presented at the upper and bottom images of the seventh step in Fig.~\ref{fig1}(a).

\item \textbf{Text Location} -- In scene text recognition, the text detection model might not crop a text box properly. That is, many scene texts would not be well-aligned to the center of the text box. To account for this, the foreground text is located at the background image with random margins. The margin is randomly set from 1 to 4 points in our synthetic dataset.

\item \textbf{Noise} - In our final rendering step, Gaussian blur is applied to the text image as shown in the ninth step of Fig.~\ref{fig1}(a). 
\end{enumerate}

In our synthetic scene text data, all the synthesized images are preprocessed by following these steps. The goal is to produce scene text images similar to the real-world ones. To evidence the effectiveness of these rendering tricks, we also conduct experiments to address these issues. The detailed results are shown in Secton \ref{exp}.

\subsection{Real-world Traditional Chinese Scene Text Dataset: TC-STR 7k-word} \label{tc-str}

It is worth repeating that the lack of public authentic Traditional Chinese scene text datasets makes evaluation more difficult. To overcome this problem, inspired by the {\tt IIIT 5K-word} dataset \cite{iiit}, we create a real-world Traditional Chinese scene text recognition dataset ({\tt TC-STR 7k-word}). Our {\tt TC-STR 7k-word} dataset collects about 1,554 images from Google image search to produce 7,543 cropped text images. To increase the diversity in our collected scene text images, we search for images under different scenarios and query keywords. Since the collected scene text images are to be used in evaluating text recognition performance, we manually crop text from the collected images and assign a label to each cropped text box. 

To optimize the ease of use of the {\tt TC-STR 7k-word}, we also split the dataset into training and testing sets by considering the distribution of characters. That is, we balance the distribution of each character between training and testing sets, making sure each character in the testing set is also found in the training set. This data splitting strategy produces a training set of 3,837 text images ({\tt TC-STR-train}) and a testing set of 3,706 images ({\tt TC-STR-test}) in our {\tt TC-STR 7k-word} dataset.

\section{Experiments}
\label{exp}

\subsection{Experimental Settings} \label{implementation_detail}
Our experiments focus on the effectiveness of the synthetic data generated by our proposed synthetic scene text engine. Several synthetic datasets are generated under different settings of the synthetic scene text engine, and the real-world scene text data, {\tt TC-STR-test}, are used to evaluate recognition performance in all experiments. For image preprocessing, all scene text images are converted to grayscale and resized to $32 \times 100$ without aspect ratio preservation.

We choose the TPS-ResNet-BiLSTM-CTC framework as our base model, respectively setting the number of fiducial points of TPS, the number of output channels of ResNet, and the size of the BiLSTM hidden state to 20, 512, and 256. For the optimizer of the learning model, we used AdaDelta \cite{adadelta} with a learning rate of $lr=1.0$ and a decay rate of $\rho=0.95$. Gradient clipping is set to magnitude 5. Models are trained with a batch size of 192 images. We validate the model after every 2,000 iterations and chose the model with the highest accuracy based on a maximum of 300K iterations.

\subsection{Trained from scratch with {\tt TC-STR-train}} \label{train-on-tcstr}

As a comparison for synthetic datasets, we use the real-world data to establish the baseline for model training. Thus, our baseline model only uses the {\tt TC-STR-train} for training (3,251 for training and 586 for validation) and is evaluated on {\tt TC-STR-test}. However, the baseline model only can achieve 4.10\% accuracy since the fonts, backgrounds, text sizes, and distortions in {\tt TC-STR 7k-word} are quite diverse.

Data augmentation is a common strategy to increase training data diversity. In our experiments, we adopted the data augmentation methods proposed in \cite{data-aug}, which achieve significant improvements on several public benchmark datasets \cite{iiit,ic13,ic15}. We scaled the original images with seven different sizes and generated augmented images with respective distortion, stretch, and perspective factors of 24, 24, and 6. Thus, the number of total images for model learning is $(1+(24+24+6)\times 7)\times 3,251=1,232,129$. 

With the augmented images, the accuracy could be improved significantly from 4.10\% to 8.82\% as shown in the second row of TABLE~\ref{table-all-results}. However, the improvement is still not adequate for practical applications. To address this issue, we use the synthetic data to improve performance in the following experiments.

\begin{figure}[t]
\begin{tikzpicture}
\begin{axis}[
    title=,
    xlabel={Number of data [$\times 1,076,764$]},
    ylabel={Test accuracy [\%]},
    xmin=0, xmax=25,
    ymin=76, ymax=88,
    xtick={0,1,5,10,15,20,25},
    ytick={74, 76, 78, 80, 82, 84, 86, 88, 90},
    legend pos=north west,
    ymajorgrids=true,
    grid style=dashed,
]

\addplot[
    color=blue,
    mark=o,
    ]
    coordinates {
    (1,78.44)(2,78.44)(3,80.14)(4,82.95)(5,82.46)(10,82.87)(15,84.11)(20,83.24)(25,83.89)
    };
    \legend{simple background}
    
\end{axis}
\end{tikzpicture}
\caption{Results on different data sizes of synthetic data with simple backgrounds.}
\label{fig-simple-bg}
\end{figure}
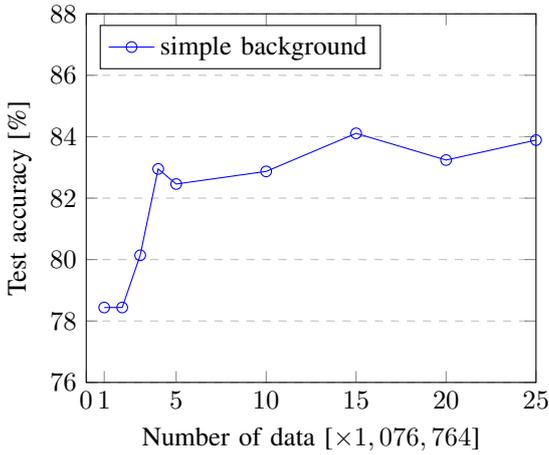

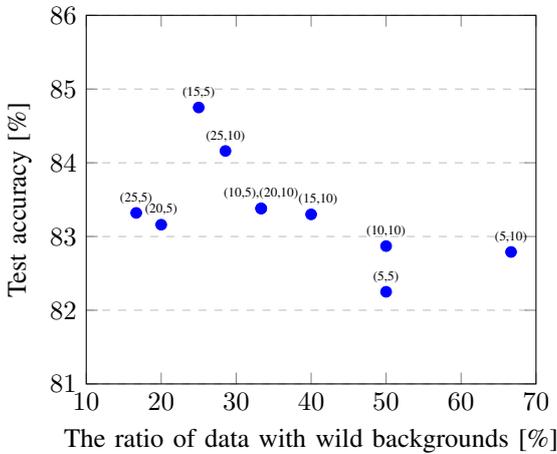
\begin{figure}[t]
\begin{tikzpicture}
\begin{axis}[%
    title=,
    xlabel={The ratio of data with wild backgrounds [\%]},
    ylabel={Test accuracy [\%]},
    xmin=10, xmax=70,
    ymin=81, ymax=86,
    xtick={0,10,20,30,40,50,60,70},
    ytick={81,82,83,84,85,86,87},
    ymajorgrids=true,
    grid style=dashed]
\addplot[
    scatter/classes={a={blue}},
    scatter, mark=*, only marks, 
    scatter src=explicit symbolic,
    nodes near coords*={\Label},
    visualization depends on={value \thisrow{label} \as \Label},
    every node near coord/.append style={font=\tiny}
] table [meta=class] {
x y class label
50.00 82.25 a (5,5)
33.33 83.38 a (10,5),(20,10)
25.00 84.75 a (15,5)
20.00 83.16 a (20,5)
16.67 83.32 a (25,5)
66.67 82.79 a (5,10)
50.00 82.87 a (10,10)
40.00 83.30 a ~~~(15,10)
33.33 83.38 a ~
28.57 84.16 a (25,10)
    };
\end{axis}
\end{tikzpicture}
\caption{Results on the different ratios of wild background images to the whole synthetic data. ($n_s$, $n_w$) represents that a training set contains $n_s$ and $n_w$ units of simple and wild background images respectively.}
\label{fig-wild-ratio}
\end{figure}

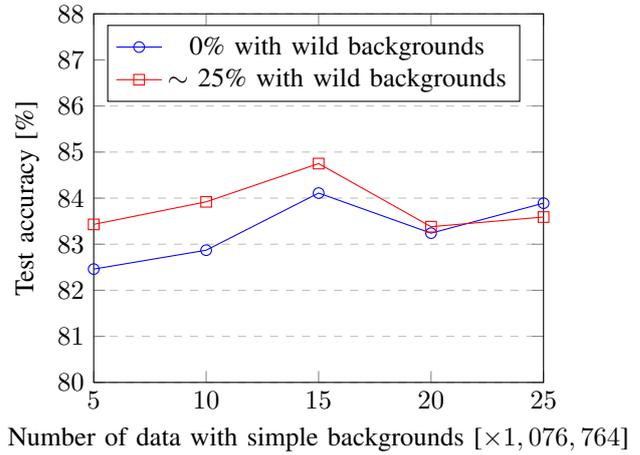
\begin{figure}[t]
\begin{tikzpicture}
\begin{axis}[
    title=,
    xlabel={Number of data with simple backgrounds [$\times 1,076,764$]},
    ylabel={Test accuracy [\%]},
    xmin=5, xmax=25,
    ymin=80, ymax=88,
    xtick={5,10,15,20,25},
    ytick={80, 81, 82, 83, 84, 85, 86, 87, 88},
    legend pos=north west,
    ymajorgrids=true,
    grid style=dashed,
]

\addplot[
    color=blue,
    mark=o,
    ]
    coordinates {
    (5,82.46)(10,82.87)(15,84.11)(20,83.24)(25,83.89)
    };
    \addlegendentry{0\% with wild backgrounds}

\addplot[
    color=red,
    mark=square,
    ]
    coordinates {
    (5,83.43)(10,83.92)(15,84.75)(20,83.38)(25,83.59)
    };
    \addlegendentry{$\sim$ 25\% with wild backgrounds}

\end{axis}
\end{tikzpicture}
\caption{Results on different data sizes of synthetic data by a fixed ratio ($\sim$25\%) of wild backgrounds to the whole synthetic data.}
\label{fig-add-wild-bg}
\end{figure}

\begin{table*}[t]
\caption{Results tested on {\tt TC-STR-test} after fine-tuning with the pre-trained model learned from our synthetic data with mixed backgrounds.}
\begin{center}
\begin{tabular}{lr|lr|r}
\hline
\multicolumn{2}{c|}{\textbf{Fine-tuning Training Data}} & \multicolumn{2}{c|}{\textbf{Fine-tuning Validation Data}} & \multicolumn{1}{c}{\textbf{Test}} \\
\multicolumn{1}{c}{\textbf{name}} & \multicolumn{1}{c|}{\textbf{\# of data}} & \multicolumn{1}{c}{\textbf{name}} & \multicolumn{1}{c|}{\textbf{\# of data}} &  \multicolumn{1}{c}{\textbf{Accuracy}} \\
\hline
Subset of {\tt TC-STR-train} & 3,251 & Subset of {\tt TC-STR-train} & 586 & 89.64\% \\
Subset of {\tt TC-STR-train} + Augmentation \cite{data-aug} & 1,232,129 & Subset of {\tt TC-STR-train} & 586 & 89.99\% \\
\hline
\end{tabular}
\label{table-finetune}
\end{center}
\end{table*}

\begin{table}[t]
\caption{Results on the ablations of background diversity, font diversity, scene diversity and word diversity.}
\begin{center}
\begin{tabular}{l|rr}
\hline
\multicolumn{1}{c|}{\textbf{Training Data}} & \multicolumn{2}{c}{\textbf{Test Accuracy}} \\
\hline
Our Synthetic Data with Mixed BGs &  84.75\% & \\
~~w/o background diversity & 8.82\% & (-75.93\%) \\
~~w/o font diversity & 62.09\% & (-22.66\%) \\
~~w/o scene diversity & 81.14\% & (-3.61\%) \\
~~w/o word diversity & 78.66\%& (-6.09\%) \\
\hline
\end{tabular}
\end{center}
\label{table-ablation}
\end{table}

\subsection{Trained from our proposed synthetic text data} \label{train-on-synth}

We conducted synthetic training experiments under different settings of the synthetic scene text engine. We trained the model only using synthetic text data and validated it on {\tt TC-STR-train} to prevent the model from over-fitting on the synthetic text data in the following experiments. 

The experiments described in this section were conducted with two purposes. First, we explore the sensitivity to synthetic data size and seek to identify a proper amount of synthetic data for our Traditional Chinese scene text recognition task. To this end, we gradually increased the synthetic text data with simple backgrounds, and evaluate the performance on {\tt TC-STR-test}. The results are shown in Fig.~\ref{fig-simple-bg}, with the x-axis representing the synthetic data size. Each unit stands for 1,076,764 images according to the size of the lexicons in our synthetic scene text engine. Performance is found to significantly improve as we initially increase the synthetic data size and tends to plateau after more than four units of synthetic data are added. In this experiment, we achieve the highest accuracy 84.11\% by using 15 units of synthetic data (see row 4 in TABLE~\ref{table-all-results}). This shows that our proposed synthetic datasets can make up for the lack of manually labeled data. 

We also investigate the effect of using wild backgrounds by comparing them with images using only simple backgrounds. Different ratios of synthetic data with wild backgrounds are considered in this experiment. In the following, we use the notation ($n_s$,$n_w$) to represent that a training set has respectively $n_s$ and $n_w$ units of simple and wild background images. We design ten combinations of training sets, including (5,5), (10,5), (15,5), (20,5), (25,5), (5,10), (10,10), (15,10), (20,10), and (25,10). The results of these training sets are presented in Fig.~\ref{fig-wild-ratio}, which shows that the recognition model achieves better accuracy roughly at a ratio of 25\%. To investigate whether 25\% is a proper ratio for mixing the wild backgrounds into our synthetic data or not, we also generate five training sets as follows: (5,2), (10,3), (15,5), (20,7), (25,8). All the ratios of wild backgrounds to the whole synthetic data are quite close to 25\%. Fig.~\ref{fig-add-wild-bg} presents the results of these five combinations and shows that generating synthetic data with a proper amount of wild background images improves performance over models trained only with simple backgrounds. 

\subsection{Ablation study} \label{ablation-study}
We have shown that our proposed synthetic scene text engine significantly improves the performance of a scene text recognition model. The use of many rendering tricks used in our synthetic data generation raises the question of the contribution of these tricks to recognition performance. To address this question, we perform ablation studies in background diversity, font diversity, scene diversity, and word diversity in our experiments. 

For the ablation of background diversity, we generate synthetic images only using black texts and white backgrounds as the training set. The results in TABLE~\ref{table-ablation} show that the unique background affected the performance significantly, and decreased accuracy. This implies that background rendering could produce synthetic scene text images similar to real-world ones.

Considering the font diversity, we used Source Han Sans as the single font without stroking in our ablation experiments. Results from TABLE~\ref{table-ablation} also show that performance is reduced without font diversity and that varied font types are beneficial for the generalization to other new fonts.

To test the ablation of scene diversity, we removed the rendering steps of skewing, distorting, and noise in our synthetic data generation. Compared with the previous ablation experiments, the influence of scene diversity is not significant from the results of TABLE~\ref{table-ablation}, possibly because there are fewer blurry images in the real world, such as the images of {\tt TC-STR 7k-word}.

For the ablation of word diversity, we removed 9/10 words from the dictionary when generating synthetic images. The results again suggest that the diversity of the character set will affect recognition performance.

\subsection{Trained and validated from our proposed synthetic text data} \label{only-synth}

In previous sections, we use {\tt TC-STR-train} or its subset as the validation set in our training procedure. However, the labeled scene text images of interest might not be easily available. Thus, we conduct an experiment which only uses synthetic data in the training procedure.

The recognition is learned by 20 units of synthetic data with mixed backgrounds (20 $\times 1,076,764$ images) as the training set and 6,000 synthetic images with mixed backgrounds as the validation set. Evaluating the learned model on {\tt TC-STR-test}, we can achieve 83.51\% accuracy as shown in the sixth row of TABLE~\ref{table-all-results}. The result shows the learned model still performs well only using synthetic images, which also suggests our synthetic engine can generate informative images similar to real ones. 

\subsection{Fine-tuning with our pre-trained model} \label{finetune}

In Section \ref{train-on-tcstr}, we show that the recognition model trained from scratch with {\tt TC-STR-train} did not perform well on the {\tt TC-STR-test}, due to the lack of training data. To overcome this problem, pre-trained models are widely used as a good initialization for deep neural networks. In our fine-tuning procedure, we first obtain the pre-trained model learned from 20 units of synthetic data with mixed backgrounds, then train the model with the subset of {\tt TC-STR-train}. The fine-tuned model achieves 89.64\% accuracy as shown in TABLE~\ref{table-finetune}, outperforming the pre-trained model (84.75\%). The model has learned the characteristics of real-world texts after pre-training, capturing more details from a small amount of labeled data. 

Besides, we also apply data augmentation to the fine-tuning procedure. It can be easily observed that improvement of recognition with data augmentation is relatively small from TABLE~\ref{table-all-results}. The possible reason is that we has generated many synthetic images for pre-trained model. Thus, the effect of augmentation in the fine-tuning procedure is not significant.

\section{Conclusion}
\label{conclusion}

Addressing the lack of a public Traditional Chinese scene text dataset, we introduce our manually labeled real-world {\tt TC-STR 7k-word} dataset offering a benchmark dataset for further research. In addition to the proposed real-world dataset, we present a framework of a Traditional Chinese synthetic data engine using the synthetic data to improve performance. In our ablation studies, we conclude that background rendering is the most critical step for generating synthetic scene text images. A model pre-trained using our proposed synthetic dataset achieves optimal accuracy (89.64\%) after fine-tuning and evaluation on the {\tt TC-STR 7k-word} dataset.


\textbf{Acknowledgment} This work was supported in part by the E.SUN Financial Holding CO., LTD. of Taiwan and the Ministry of Science and Technology of Taiwan under Grants MOST 108-2221-E-017-008-MY3.




%

\end{document}